\documentclass{article}
\pdfoutput=1

\usepackage{arxiv}

\usepackage[T1]{fontenc}    
\usepackage{hyperref}       
\usepackage{url}            
\usepackage{booktabs}       
\usepackage{amsfonts}       
\usepackage{nicefrac}       
\usepackage{microtype}      
\usepackage{lipsum}

\usepackage{amsmath}        
\usepackage{graphicx}
\usepackage{braket}         
\usepackage{preamble}       
\usepackage{subcaption}     
\usepackage{float}          

\usepackage{cite}

\title{An Approach to Symbolic Regression Using Feyn}
\author{
  Kevin Ren\'{e} Brol\o{}s \\
  Abzu\\
  Copenhagen, Denmark \\
  \texttt{kevin.broloes@abzu.ai} \\
  \And
  Meera Vieira Machado \\
  Abzu\\
  Copenhagen, Denmark \\
  \texttt{meera.machado@abzu.ai} \\
  \And
  Chris Cave \\
  Abzu \\
  Copenhagen, Denmark \\
  \texttt{chris.cave@abzu.ai} \\
  \And
  Jaan Kasak \\
  Abzu \\
  Copenhagen, Denmark \\
  \texttt{jaan.kasak@abzu.ai} \\
  \And
  Valdemar Stentoft-Hansen \\
  Abzu \\
  Copenhagen, Denmark \\
  \texttt{valdemar.stentoft@abzu.ai} \\
  \And
  Victor Galindo Batanero \\
  Abzu Barcelona\\
  Barcelona, Spain \\
  \texttt{victor.galindo@abzu.ai} \\
  \And
  Tom Jelen \\
  Abzu Barcelona \\
  Barcelona, Spain \\
  \texttt{tom.jelen@abzu.ai} \\
   \And
  Casper Wilstrup \\
  Abzu\\
  Copenhagen, Denmark \\
  \texttt{casper.wilstrup@abzu.ai} \\
  }

\begin{document}
\maketitle

\begin{abstract}

In this article we introduce the supervised machine learning tool called \texttt{Feyn}. The simulation engine that powers this tool is called the \texttt{QLattice}. The \texttt{QLattice} is a supervised machine learning tool inspired by Richard Feynman's path integral formulation, that explores many potential models that solves a given problem. It formulates these models as graphs that can be interpreted as mathematical equations, allowing the user to completely decide on the trade-off between interpretability, complexity and model performance. 

We touch briefly upon the inner workings of the \texttt{QLattice}, and show how to apply the python package, \texttt{Feyn}, to scientific problems. We show how it differs from  traditional machine learning approaches, what it has in common with them, as well as some of its commonalities with symbolic regression. We describe the benefits of this approach as opposed to black box models.

To illustrate this, we go through an investigative workflow using a basic data set and show how the \texttt{QLattice} can help you reason about the relationships between your features and do data discovery.

\end{abstract}


\newpage

\section{Introduction}

The \texttt{QLattice} is a supervised machine learning tool for symbolic regression. It composes functions together to build graphs of mathematical models, adding interactions between the features in your data set and your target variable. The functions vary from elementary ones such as addition, multiplication, squaring, to more complex ones such as natural logarithm, exponential and tanh. The \texttt{QLattice} is a technology developed by \texttt{Abzu} that is inspired by Richard Feynman's \href{https://en.wikipedia.org/wiki/Path_integral_formulation}{path integral formulation}. Symbolic regression, which is gaining a lot of momentum these days\cite{Dabhi2011, Vladislavleva2009, schmidt2009, cranmer2020, navarro2020, kim2020,cranmer2019}. Other symbolic regression methods inspired by physics are also getting traction\cite{udrescu2020, tailin2019, liu2020}. Overall, symbolic regression approaches are showing signs of keeping a high performance, while still maintaining generalisability\cite{orzechowski2018}, which separates it from other popular graph-like models such as decision trees or random forests\cite{Breiman2001}.

These mathematical models are represented by graphs that are fitted using \texttt{Feyn}, a Python module that allows you to interface with the \texttt{QLattice}. Each graph is generated from a probability distribution, that is tuned over time. Initially, this probability distribution is uniform.

So how do you tune this probability distribution?

This follows an iterative process. You'll fit thousands of graphs representing different function compositions, suggested to you by the \texttt{QLattice}. You'll then select the best ones, and update the \texttt{QLattice} with them. This helps the \texttt{QLattice} converge and shape the probability distribution towards better solutions.

The space of all possible models is very large, and without guidance the \texttt{QLattice} will take a very long time to converge. This requires you to be specific on what you want to investigate in your data set. Typically, this means restricting the types of graphs that the \texttt{QLattice} will produce. The \texttt{QLattice} is useful when you want insights and investigate relationships between your features, rather than focus exclusively on predictive power. This process is what we'll go through in this paper.

\newpage

\section{A quick primer on the QLattice and Feyn}
\subsection{The QLattice in a nutshell}

The \texttt{QLattice} is an environment to simulate discrete paths from multiple inputs to an output. It does this in a finite multi-dimensional lattice-space. This is where the inspiration from Feynman's path integral comes in. The \texttt{QLattice} simulates inputs as originating anywhere, taking a (short) path through the lattice space, before emerging to an output. If you imagine doing this thousands of times, until a solid path has been shaped, you'll experience a convergence to the most likely or, in this case, most useful path to explain the problem you're trying to model. Along the path that we take, we'll randomly sample from a selection of "interactions" -- functions that transform the inputs to a new output. These can vary from elementary ones such as addition, multiplication, squaring, to more complex ones such as natural logarithm, exponential and tanh.

We determine the interactions based on probabilities, guided by repeated reinforcement of the best solutions provided by the \texttt{QLattice}, as you fit the thousands of paths (models), that are discovered. During repeated reinforcement, another phenomenon that happens, is that islands in the \texttt{QLattice} space starts to form, each with their own independent evolution. This narrows the search space, and gives way to many separate evolutionary spaces. Another benefit to this process, is that the user helps decide which models are useful, and through that which paths will be reinforced. The user also decides how to constrain the decision space, giving the user full control over the shapes the models will be taking.

Altogether, this approach has some benefits, such as:
\begin{itemize}
    \item there are far fewer nodes and connections (and we also call the nodes `interactions`)
    \item there are functions you wouldn't normally see in a neural network (such as `gaussian`, `inverse` and `multiply`)
    \item the graphs are more inspectable, simpler and less prone to overfitting
    \item the graphs allow you to interpret them as a mathematical formula, allowing you to reason about the consequences of your hypothesis.
\end{itemize}

\subsection{Introducing the QGraph}
So let's talk more about how it does this. If you recall the paths with interactions above, these are represented by unidirectional, acyclic graphs. But before we can start getting to these graphs, we need to consider the concept of the \texttt{QGraph}. The \texttt{QGraph} is a generator, that represents a subset of the possibilities of the \texttt{QLattice}. This subset is defined by you by deciding on things like: Is it a classification or regression problem, which features do I want to learn about, what interactions do I want, how deep will I allow my graphs to be, and other such constraints. 

So instead of producing a graph, the \texttt{QLattice} produces a \texttt{QGraph}. In an ideal world, the \texttt{QGraph} would contain the infinite list of graphs that matches your constraints. In the practical world, it is limited to a list of a few thousand graphs at a time, and will instead continuously replenish this list, and discard the worst graphs based on your evaluation criteria. The criteria available are common ones, such as cross-entropy, RMSE, Bayesian Information Criterion, Akaike Information Criterion\cite{Akaike1974}, etc.

\subsection{Feyn}

\texttt{Feyn}\cite{feyn} is the python module for interfacing with the \texttt{QLattice}, through \texttt{QGraphs}. 

In \texttt{Feyn}, like other machine learning tools, a model is fitted using a variation of backpropagation\cite{backprop} on the dataset. The similarities end quickly, however, as the \texttt{QLattice} helps you try many different alternative models, and is typically used with the purpose of understanding and explaining your problem, or answering the questions you wish to know more about.

On top of this, we've added some quality-of-life features for explainability and inspection, as well as automatic encoding of input features. This means that normalisation is not necessary, and you should also not one-hot encode categorical variables, as it gets automatically handled and represented as a single feature that's easier to understand.

Throughout the next few sections you'll see plenty of examples of what these models look like, but let's go a little into how it's used.

\subsection{Privacy and transfer of information}

It's worth noting there, that the information sent to the \texttt{QLattice} only consists of:
\begin{itemize}
    \item The token and \texttt{QLattice} identifiers when you connect.
    \item The names of your columns (text strings)
    \item The graphs you want to reinforce, containing column names, function names (interactions), along with the unweighted edges in the graph that connect inputs to functions, functions to functions and functions to outputs.
\end{itemize}

None of your data is at any point exchanged and does not leave your machine.

\subsection{Features and functions}

We should also talk a bit more about:
\begin{itemize}
\item the \textit{semantic types}
\item the \textit{interactions} in a graph
\end{itemize}

\subsubsection{Semantic types}

In order to create your \texttt{QGraph}, you first need to tell the \texttt{QLattice} what type of data the inputs and output are. This is a way of telling it what types of data it should expect and from that how to behave. We call these \textit{semantic types} or \textit{stypes}. There are two \textit{stypes}:

\begin{itemize}
\item \textbf{Numerical}: This is for features that are continuous in nature such as: height, number of rooms, latitude and longitude etc.
\item \textbf{Categorical}: This is for features that are discrete in nature such as: neighbourhoods, room type etc.
\end{itemize}

If no type is assigned to the feature, the numerical semantic type is assumed. This means you will only need to assign an \textit{stype} to each categorical feature. This is how the graph knows to do the automatic encoding of the categories.

\subsubsection{Interaction}

\textit{Interactions} are the basic computation units of each model. They take in data, transform it and then emit it out to be used in the next \textit{interaction}. Here are the current possible interactions:

\begin{longtable}[width=0.6\textwidth]{@{}
  >{\raggedright\arraybackslash}p{(\columnwidth - 2\tabcolsep) * \real{0.25}}
  >{\raggedright\arraybackslash}p{(\columnwidth - 2\tabcolsep) * \real{0.25}}@{}}
\toprule
Type & Function \\ \addlinespace
\midrule
\endhead
Addition & $a + b$ \\ \addlinespace
Multiply & $a * b$ \\ \addlinespace
Squared & $a * a$ \\ \addlinespace
Linear & $a*weight+bias$ \\ \addlinespace
Tanh & $tanh(a)$ \\ \addlinespace
Single-legged Gaussian & $e^{-a^{2}}$ \\ \addlinespace
Double-legged Gaussian & $e^{-(a^{2} + b^{2})}$ \\ \addlinespace
Exponential & $exp(a)$ \\ \addlinespace
Logarithmic & $log(a)$ \\ \addlinespace
Inverse & $\frac{1}{a}$ \\ \addlinespace
\bottomrule
\end{longtable}

\newpage

\section{Connecting the scientific method and the QLattice}
At \texttt{Abzu}, we put the scientific method at the core of our workflow, and use this approach to understand data, generate hypotheses and validate the findings against the actual observations.

\begin{figure}
\centering
\includegraphics[width=0.6\textwidth]{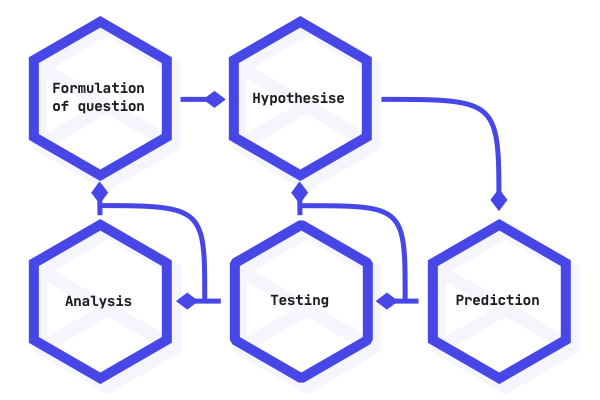}
\caption{The Scientific Method as an iterative process}
\label{fig:scientific_method}
\end{figure}

The \href{https://en.wikipedia.org/wiki/Scientific_method\#Process}{scientific method} can be summed up in short with a diagram (Figure \ref{fig:scientific_method})

In summary, using \texttt{Feyn} and the \texttt{QLattice} would look something like this:

\begin{longtable}[width=0.95\textwidth]{@{}
  >{\raggedright\arraybackslash}p{(\columnwidth - 2\tabcolsep) * \real{0.25}}
  >{\raggedright\arraybackslash}p{(\columnwidth - 2\tabcolsep) * \real{0.75}}@{}}
\toprule
Scientific Method & In practice \\ \addlinespace
\midrule
\endhead
Formulation of question & Determine what your research question is. \\ \addlinespace
Hypothesise & Use the \texttt{QLattice} to generate possible hypotheses in the form of graphs that answer your question. \\ \addlinespace
Prediction & Determine how your graph predicts existing data and plot what the graph tells you about the things you haven't seen before. \\ \addlinespace
Testing & Take your hypothesis to the lab, real life, or validate it on a holdout set, depending on your needs and capabilities. \\ \addlinespace
Analysis & Accept or reject a hypothesis based on the results you get from your experiments. \\ \addlinespace
\bottomrule
\end{longtable}

\newpage

\section{Asking the right question to the QLattice}
Following from the \href{introduction.md}{workflow overview}, we'll go
through the first two steps - namely making observations and posing
interesting questions.

\hypertarget{make-observations}{%
\subsection{Make observations}\label{make-observations}}

Let's take the following simple dataset. This is the well-known
\href{https://goo.gl/U2Uwz2}{UCI ML Breast Cancer Wisconsin (Diagnostic)
dataset}\cite{breast-cancer-dataset}. We load it through the commonly used sklearn\cite{scikit-learn} machine learning package.

Features are computed from a digitized image of a fine needle aspiration
(FNA) of a breast mass. They describe characteristics of the cell nuclei
present in the image.

The defined target variable is the diagnosis, as `Malignant' or
`Benign'.

\begin{lstlisting}[numbers=left, language=python, stringstyle=\color{strings}, commentstyle=\color{comments}, morekeywords={import,from,for,in,as,range,if,else,elif}, keywordstyle=\color{keywords}\bfseries, basicstyle=\small]
import sklearn.datasets
import pandas as pd

breast_cancer = sklearn.datasets.load_breast_cancer()
input_columns = breast_cancer.feature_names

# Load into a pandas dataframe
data = pd.DataFrame(breast_cancer.data, columns=input_columns)
data['target'] = pd.Series(breast_cancer.target)
\end{lstlisting}

\hypertarget{preparation}{%
\subsubsection{Preparation}\label{preparation}}

This dataset comes prepared already so you don't have to do quite as
much. The \texttt{QLattice} workflow typically starts after data preparation, however, it
is worth mentioning that with the \texttt{QLattice},
you \emph{don't} need to do any normalization of
\href{../essentials/inputs.md}{input features}, and we have an input
that explicitly handles categorical variables without the need for
one-hot encoding.

\hypertarget{the-features}{%
\subsubsection{The features}\label{the-features}}

\begin{figure}
\centering
\includegraphics[height=0.9\textheight]{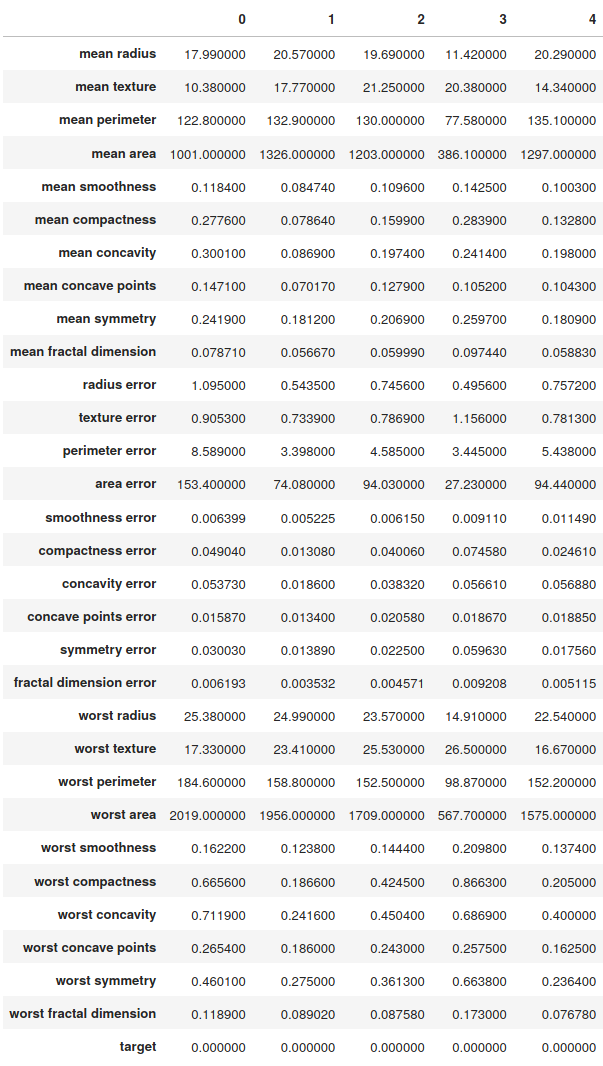}
\caption{The first five data samples in our dataframe, oriented column-wise}
\label{fig:data_head_breast_cancer}
\end{figure}

In Figure \ref{fig:data_head_breast_cancer} we have printed the head of the dataframe so we can see what we're working
with.

\begin{lstlisting}[numbers=left, language=python, stringstyle=\color{strings}, commentstyle=\color{comments}, morekeywords={import,from,for,in,as,range,if,else,elif}, keywordstyle=\color{keywords}\bfseries, basicstyle=\small]
data.head().T
\end{lstlisting}

\hypertarget{asking-questions}{%
\subsection{Asking Questions}\label{asking-questions}}

We'll now go through some examples of interesting questions you could
pose to this dataset. This comes down to your domain expertise and
figuring out what you want to learn about, such as: 

\begin{itemize}
	\item What is your measurable - also known as the target variable. This forms the basis of our questions as the variable you want to explain.
	\item What are you trying to learn?
    \begin{enumerate}
    	\item \textit{Example: What is evidence for a malignant or benign tumor?}
        \item \textit{Example: We have a specific feature we suspect is evidence for malignant tumors based on previous studies. We could ask whether this feature relates to the diagnosis, and if so, how?}
    \end{enumerate} 
\end{itemize}

If you don't have specific knowledge, but are trying to learn things
about the problem domain, maybe you'll ask more general questions:

\begin{itemize}
	\item Is there a single feature that captures a large part of the signal?
	\begin{enumerate}
	    \item If it does, how does it capture it - i.e.~linearly or non-linearly?
    	\item Are there some features that tend to explain the \textbf{same} signal?
        \item Are there some features that tend to explain \textbf{different} parts of the signal?
    \end{enumerate}
    \item If there's one feature that captures part of the signal, does it then combine with another to capture even more?
    \begin{enumerate}
	    \item How do the features relate to explain the measurable?
    	\item Does it make sense from a domain perspective?
        \item Would that give way for new questions to ask?
    \end{enumerate}
    \item Or whichever other question you might have.
\end{itemize}

Other datasets might have multiple interesting targets to measure
against, so remember to choose the measurable(s) that will be the
strongest indicator of the questions you're asking.

This is an iterative process, so you can always go back and change this as you learn more.

\hypertarget{on-to-hypotheses}{%
\subsubsection{On to hypotheses}\label{on-to-hypotheses}}

As your next step, you will be using the \texttt{QLattice}
to pose questions and formulate hypotheses as answers to some of these
questions.

\section{Formulating hypotheses}
From the previous page, we started asking questions about our data. Our
overarching question is: \textit{`What is evidence for a malignant or
benign tumor?'}. Throughout the guides we will pose more specific
questions and hypotheses which will help us in answering this main
question.

We now want to use the \texttt{QLattice} to help us
find good hypotheses to these questions.

\hypertarget{from-questions-to-hypotheses}{%
\subsection{From questions to
hypotheses}\label{from-questions-to-hypotheses}}

First we need to translate the question into a
\texttt{QGraph}. Take for instance the question
\textit{`Is area indicative of a tumor being benign or malignant?'}, the
corresponding \texttt{QGraph} would look like the
following:

\begin{lstlisting}[numbers=left, language=python, stringstyle=\color{strings}, commentstyle=\color{comments}, morekeywords={import,from,for,in,as,range,if,else,elif}, keywordstyle=\color{keywords}\bfseries, basicstyle=\small]
import sklearn.datasets
import pandas as pd
import feyn

from sklearn.model_selection import train_test_split

breast_cancer = sklearn.datasets.load_breast_cancer()
input_columns = breast_cancer.feature_names

# Load into a pandas dataframe
data = pd.DataFrame(breast_cancer.data, columns=input_columns)
data['target'] = pd.Series(breast_cancer.target)

# Split into train, validation and holdout
train, valid = train_test_split(data, test_size = 0.4, 
                                stratify = data['target'], random_state = 42)
valid, holdo = train_test_split(valid, test_size = 0.5, 
                                stratify = valid['target'], random_state = 42)
\end{lstlisting}

Throughout our investigation we will fix a
\textit{train/validation/holdout} split. We will generate hypotheses based
on the \textit{train} set, analyse and select them on
the \textit{validation} set. Lastly, when we settle on a
hypothesis we are satisfied with we will test it on the
\textit{holdout} set.

\begin{lstlisting}[numbers=left, language=python, stringstyle=\color{strings}, commentstyle=\color{comments}, morekeywords={import,from,for,in,as,range,if,else,elif}, keywordstyle=\color{keywords}\bfseries, basicstyle=\small]
# Pose question to QGraph
ql = feyn.QLattice()
qgraph = ql.get_classifier(['mean area'], 'target', max_depth=1)
\end{lstlisting}

We have told the \texttt{QGraph} to take
\textbf{mean area} as an input feature and to map it to
the output, \textbf{target}. The
\textit{max\_depth = 1} ensures that the input feature
enters a single interaction cell as we will soon show.

The \texttt{QGraph} is a list of potential hypotheses
to the question. At the moment, the \texttt{QGraph} has
no knowledge of the data. So any hypothesis in the
\texttt{QGraph} will be random.

In order to change that, we need to fit the
\texttt{QGraph} to the data.

\begin{lstlisting}[numbers=left, language=python, stringstyle=\color{strings}, commentstyle=\color{comments}, morekeywords={import,from,for,in,as,range,if,else,elif}, keywordstyle=\color{keywords}\bfseries, basicstyle=\small]
for _ in range(20):
    qgraph.fit(train, threads=4)
    ql.update(qgraph.best())
\end{lstlisting}

We can find the best hypotheses in the \texttt{QGraph}
by calling \texttt{qgraph.best()}. This is usually a
list of 3 to 4 top hypotheses based on the closest fit to the data
i.e.~a loss function.

The \texttt{QLattice} is an environment that searches
all possible hypotheses to questions posed to it. In order to refine its
search we need to tell the \texttt{QLattice} the best
hypotheses we have seen so far. We do this by calling
\texttt{ql.update(qgraph.best())}.

Note that you can choose the number of \textit{threads} to allow for
parallel \texttt{QGraph} fitting. That will accelerate
the process.

After this fitting loop we want to take a look at what the
\texttt{QLattice} came up with.

\begin{lstlisting}[numbers=left, language=python, stringstyle=\color{strings}, commentstyle=\color{comments}, morekeywords={import,from,for,in,as,range,if,else,elif}, keywordstyle=\color{keywords}\bfseries, basicstyle=\small]
qgraph.head()
\end{lstlisting}

\begin{figure}
\centering
\includegraphics[height=0.5\textheight]{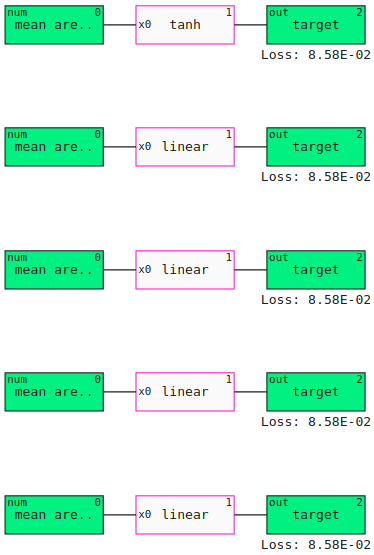}
\caption{The first five graphs of a \texttt{QGraph} filtered to always contain \textbf{mean area} and be of max depth 1}
\label{fig:qgraph_head_Q1}
\end{figure}

The output graphs in Figure \ref{fig:qgraph_head_Q1} illustrate what we said about
\textbf{mean area} entering a single interaction cell
to predict the \textbf{target} variable.

Let's increase complexity by posing another question to the
\texttt{QGraph}: \textit{`What feature could
\textbf{mean area} combine with to predict the
\textbf{target} variable?'}.

\begin{lstlisting}[numbers=left, language=python, stringstyle=\color{strings}, commentstyle=\color{comments}, morekeywords={import,from,for,in,as,range,if,else,elif}, keywordstyle=\color{keywords}\bfseries, basicstyle=\small]
qgraph = ql.get_classifier(['mean texture',
                            'mean area',
                            'mean smoothness',
                            'mean compactness',
                            'mean concavity',
                            'mean concave points',
                            'mean symmetry',
                            'mean fractal dimension'],
                           output='target',
                           max_depth=1)\
.filter(feyn.filters.Contains('mean area'))

for _ in range(50):
    qgraph.fit(train, threads=4)
    ql.update(qgraph.best())
\end{lstlisting}

We want to explore the possible features
\textbf{mean area} could combine to predict the
\textbf{target}. So we include a list of potential
ones. We choose the means because we are trying to find a simple set of
explanations to the problem. The features
\textbf{mean radius} and
\textbf{mean perimeter} were excluded since they
correlate heavily with \textbf{mean area}.

\begin{figure}
\centering
\includegraphics[height=0.70\textheight]{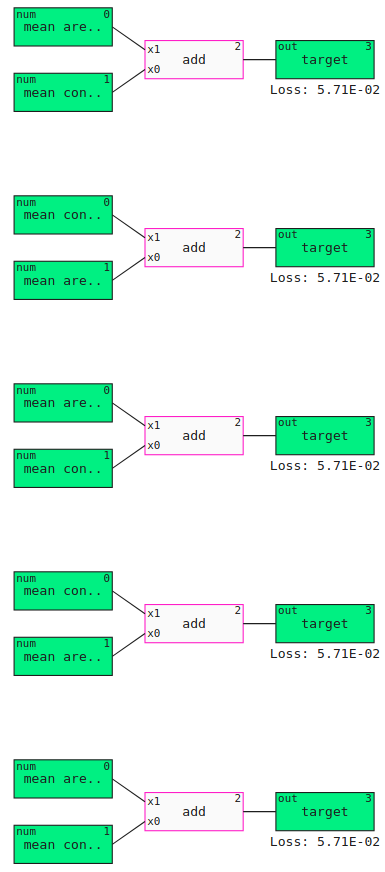}
\caption{The first five graphs of a \texttt{QGraph} filtered for 'mean area' combining with other features to predict the target}
\label{fig:qgraph_head_Q1a}
\end{figure}

We use the \textit{filters module} to ensure that only
hypotheses that correspond the question being asked are generated. The
QGraph will then only show hypotheses that satisfy the conditions
imposed by the \textit{filters}. In the case above,
\texttt{feyn.filters.Contains('mean area')} ensures
that \textbf{mean area} is included in every hypothesis
in the \texttt{QGraph} (Figure \ref{fig:qgraph_head_Q1a}).

\begin{lstlisting}[numbers=left, language=python, stringstyle=\color{strings}, commentstyle=\color{comments}, morekeywords={import,from,for,in,as,range,if,else,elif}, keywordstyle=\color{keywords}\bfseries, basicstyle=\small]
qgraph.head()
\end{lstlisting}

In the example above \textbf{mean concave points} was
the most prevalent, suggesting that it is one of the features that best
combine with \textbf{mean area}. Note how the loss has
decreased in comparison to the previous figure. This leads us to refine
our previous question: \textit{``how does
\textbf{mean area} and
\textbf{mean concave points} combine to predict the
target variable?''}.

\begin{lstlisting}[numbers=left, language=python, stringstyle=\color{strings}, commentstyle=\color{comments}, morekeywords={import,from,for,in,as,range,if,else,elif}, keywordstyle=\color{keywords}\bfseries, basicstyle=\small]
qgraph = ql.get_classifier(['mean area', 'mean concave points'], 
                            output='target', 
                            max_depth=1)

for _ in range(20):
    qgraph.fit(train, threads=4)
    ql.update(qgraph.best())
\end{lstlisting}

The next step consists in selecting a hypothesis and diving deeper into
it. The \texttt{QGraph} is sorted by loss, so
\texttt{qgraph[0]} is the hypothesis that best fit the
data passed to the \texttt{QGraph}. Most likely this
will be the hypothesis we further explore.

\newpage

\section{Analysing and selecting hypotheses}
In this section we lay the arsenal of tools to analyse and aid in the
selection of the most interesting hypotheses. Previously, we
posed the question of how \textbf{mean area} combines
with \textbf{mean concave points} to predict the target
variable. We then generated a list of hypotheses, the
\texttt{QGraph}, that could answer said question.
Lastly, we selected \texttt{qgraph[0]} as the
hypothesis to further investigate. This process is recapped below:

\begin{lstlisting}[numbers=left, language=python, stringstyle=\color{strings}, commentstyle=\color{comments}, morekeywords={import,from,for,in,as,range,if,else,elif}, keywordstyle=\color{keywords}\bfseries, basicstyle=\small]
import sklearn.datasets
import pandas as pd
import feyn
import matplotlib.pyplot as plt

from sklearn.model_selection import train_test_split

breast_cancer = sklearn.datasets.load_breast_cancer()
input_columns = breast_cancer.feature_names

# Load into a pandas dataframe
data = pd.DataFrame(breast_cancer.data, columns=input_columns)
data['target'] = pd.Series(breast_cancer.target)

# Split into train, validation and holdout
train, valid = train_test_split(data, test_size = 0.4, 
                                stratify = data['target'], random_state = 42)
valid, holdo = train_test_split(valid, test_size = 0.5, 
                                stratify = valid['target'], random_state = 42)

# Pose a question to QGraph
ql = feyn.QLattice()
qgraph = ql.get_classifier(['mean area', 'mean concave points'], output='target', max_depth=1)

for _ in range(20):
    qgraph.fit(train, threads=4)
    ql.update(qgraph.best())

hypo_mean_area_conc_points = qgraph[0]
\end{lstlisting}

\begin{figure}
\centering
\includegraphics[width=0.5\textwidth]{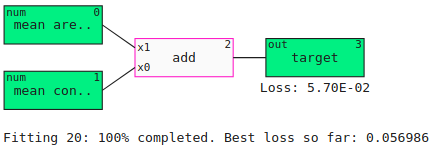}
\caption{A \texttt{feyn} graph representing hypothesis 1}
\label{fig:hypo_mean_area_conc_points}
\end{figure}

\begin{figure}
\centering
\includegraphics[width=0.75\textwidth]{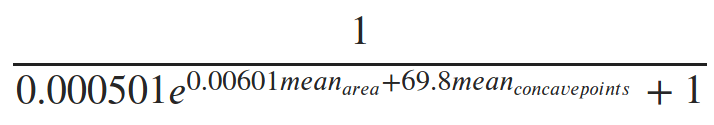}
\caption{Hypothesis 1 converted to a mathematical equation}
\label{fig:sympify}
\end{figure}

Our hypothesis (Figure \ref{fig:hypo_mean_area_conc_points}) states that the \textbf{target} variable
is a linear function of \textbf{mean area} and
\textbf{mean concave points}. We can convert this graph
to a mathematical equation using \texttt{SymPy} (Figure \ref{fig:sympify}):

\begin{lstlisting}[numbers=left, language=python, stringstyle=\color{strings}, commentstyle=\color{comments}, morekeywords={import,from,for,in,as,range,if,else,elif}, keywordstyle=\color{keywords}\bfseries, basicstyle=\small]
hypo_mean_area_conc_points.sympify(signif = 3)
\end{lstlisting}

As this is a classification problem the linear relationship is passed to
a logisitic function to obtain values between 0 and 1. This represents
the probability of a tumor being benign.

\hypertarget{analysing-a-hypothesis}{%
\subsection{Analysing a hypothesis}\label{analysing-a-hypothesis}}

\begin{figure}[hbt!]
\centering
\includegraphics[width=0.75\textwidth]{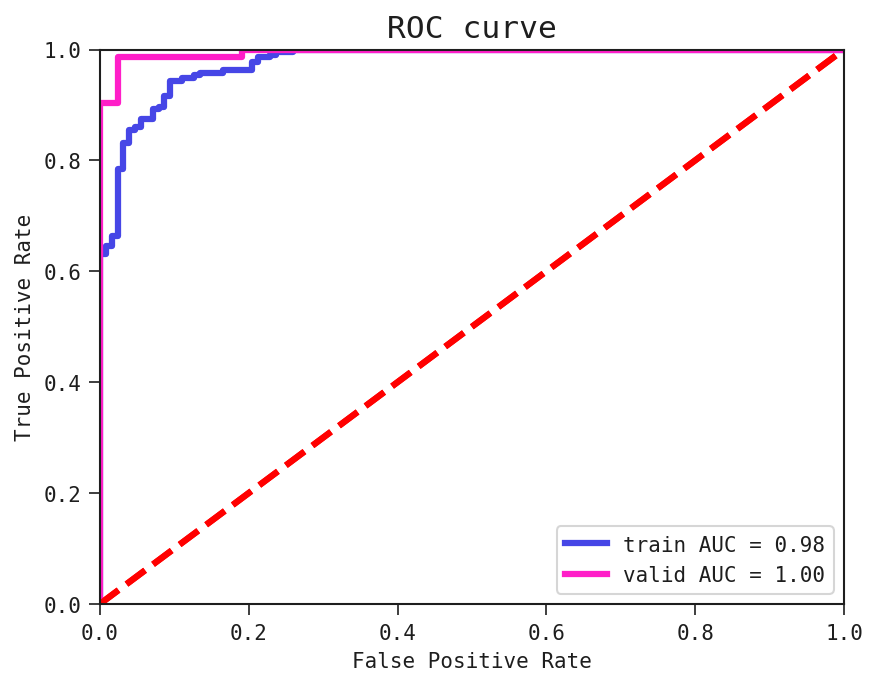}
\caption{Roc curve of our first hypothesis on the training and validation set}
\label{fig:roc_curve_hypo_mean_area_conc_points}
\end{figure}

Let's check the performance of the hypothesis above by plotting the ROC-curve (Figure \ref{fig:roc_curve_hypo_mean_area_conc_points}) on the train and validation sets. This can also tell us about overfitting, especially if their AUC scores differ significantly.

\begin{lstlisting}[numbers=left, language=python, stringstyle=\color{strings}, commentstyle=\color{comments}, morekeywords={import,from,for,in,as,range,if,else,elif}, keywordstyle=\color{keywords}\bfseries, basicstyle=\small]
hypo_mean_area_conc_points.plot_roc_curve(train, label = 'train')
hypo_mean_area_conc_points.plot_roc_curve(valid, label = 'valid')
\end{lstlisting}

The ROC curve in Figure \ref{fig:roc_curve_hypo_mean_area_conc_points} tells us that the hypothesis' predictions are not just result of random guessing. A ROC curve can be
complemented by plotting the probability scores, i.e.~the values predicted by our current hypothesis.

\begin{lstlisting}[numbers=left, language=python, stringstyle=\color{strings}, commentstyle=\color{comments}, morekeywords={import,from,for,in,as,range,if,else,elif}, keywordstyle=\color{keywords}\bfseries, basicstyle=\small]
from feyn.plots import plot_probability_scores

y_train_true = train['target'].copy()
y_train_pred = hypo_mean_area_conc_points.predict(train)
plot_probability_scores(y_train_true, y_train_pred, title='training set')
\end{lstlisting}

\begin{figure}
\centering
\includegraphics[width=0.75\textwidth]{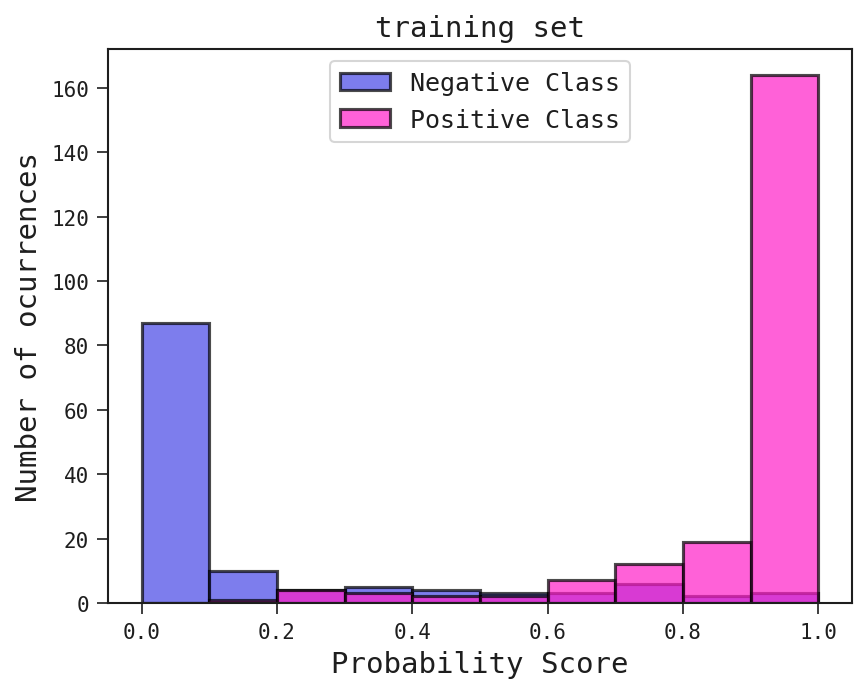}
\caption{Probability scores of our predictions}
\label{fig:probability_hypo_mean_area_conc_points}
\end{figure}

The higher the AUC score, the easier it will be to separate the negative (\textbf{target} = 0) and positive (\textbf{target} = 1) classes in the probability score plot on Figure \ref{fig:probability_hypo_mean_area_conc_points}.

We can also visualise the hypothesis with a two-dimensional partial
plot:

\begin{lstlisting}[numbers=left, language=python, stringstyle=\color{strings}, commentstyle=\color{comments}, morekeywords={import,from,for,in,as,range,if,else,elif}, keywordstyle=\color{keywords}\bfseries, basicstyle=\small]
hypo_mean_area_conc_points.plot_partial2d(train)
\end{lstlisting}

\begin{figure}
\centering
\includegraphics{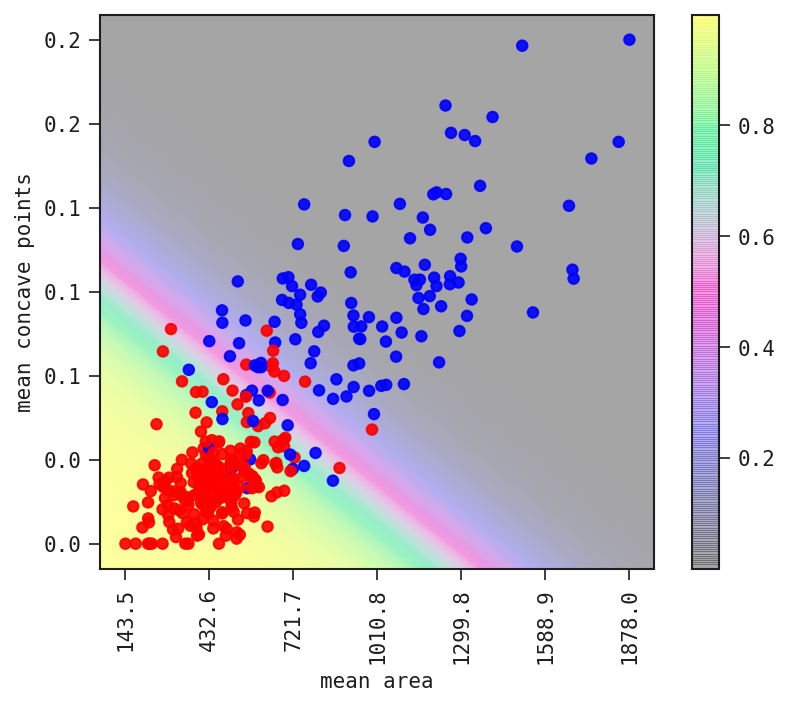}
\caption{2D Partial plot of predictions over \textbf{mean area} and \textbf{mean concave points}. Dot colors represent actual classes.}
\label{fig:hypo_mean_area_conc_points_partial2d}
\end{figure}

The red dots in Figure \ref{fig:hypo_mean_area_conc_points_partial2d} represent the positive class, benign tumors, while the blue ones represent the negative class, malignant tumors. The background colours represent what our hypothesis predicts: the yellow regions as 1, and the grey regions as 0. Since the hypothesis is a linear function it formed a straight boundary between the red and blue dots and the ROC curve above shows that this is a good separation.

The next question is how we can point to the feature values the model has difficulties classifying. This is when we use the \texttt{plot\_segmented\_loss} method.

\begin{lstlisting}[numbers=left, language=python, stringstyle=\color{strings}, commentstyle=\color{comments}, morekeywords={import,from,for,in,as,range,if,else,elif}, keywordstyle=\color{keywords}\bfseries, basicstyle=\small]
fig = plt.figure(figsize=(20,6))
ax = fig.add_subplot(121)
hypo_mean_area_conc_points.plot_segmented_loss(train, by='mean area', ax=ax)
ax = fig.add_subplot(122)
hypo_mean_area_conc_points.plot_segmented_loss(train, by='mean concave points', ax=ax)
\end{lstlisting}

\begin{figure}
\centering
\includegraphics[width=\textwidth]{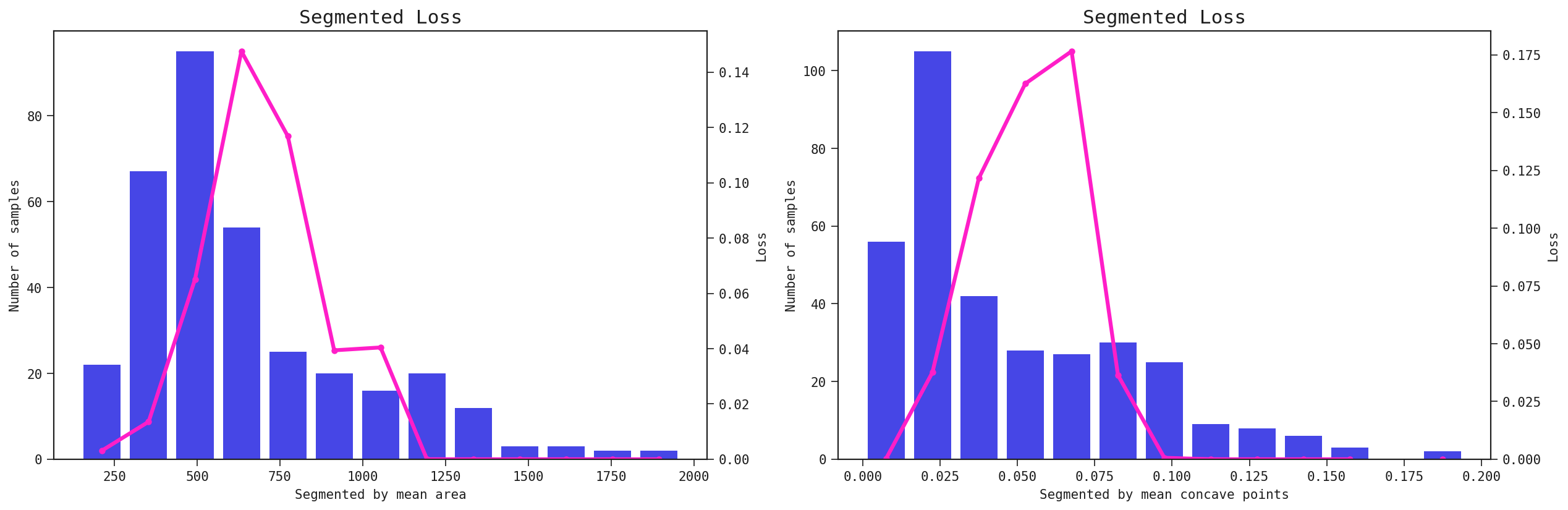}
\caption{Segmented loss by \textbf{mean area} and \textbf{mean concave points}}
\label{fig:seg_loss_mean_area_conc_points}
\end{figure}

In the left diagram of Figure \ref{fig:seg_loss_mean_area_conc_points}, the blue bars is the histogram of
\textbf{mean area}. On top of the histogram is a pink
curve where each point is the average loss across that bin. For
\textbf{mean area}, values greater than 1250 the average
loss is very close to zero. Meanwhile, the loss is much higher between
the values 500 and 750. This means the hypothesis yields predictions
closer to the observed data when \textbf{mean area}
\(> 1250\) than for \(500 <\) \textbf{mean area}
\(< 750\). We can say something similar about
\textbf{mean concave points}.

\hypertarget{concluding-remarks}{%
\subsection{Concluding remarks}\label{concluding-remarks}}

Let's go back to our overarching question: \textit{`What is evidence for a
malignant or benign tumor?'}. Throughout the sessions, we refined it to
a question on how does \textbf{mean area} and
\textbf{mean concave points} predict the
\textbf{target} variable. At last we reached the
hypothesis that the probability of having a benign tumor is the linear
function of \textbf{mean area} and
\textbf{mean concave points} shown above.

In addition to testing this hypothesis with more data, like the holdout
or newly gathered data, we can refine the question, come up with new
hypotheses, etc. In order words, our investigation of this dataset
doesn't need to stop now.

\section{Conclusion}
The hypothesis in the previous section combined
\textbf{mean area} and
\textbf{mean concave points} to differentiate between
malignant and benign tumors. However there could be other features,
other relationships to explore. So the question really comes to:

\textbf{Are we satisfied with the hypothesis we've chosen?}

A guideline to answer this question might be in assessing whether your
hypothesis points towards something we didn't know before.

If we are not satisfied, then we can go back and refine the questions
we've made based on our analysis so far. For example we could ask how
much can \textbf{mean concave points} explain the
target variable on its own or whether it can combine with other features
that do not correlate so heavily with it (check Pearson's correlation
between \textbf{mean area} and
\textbf{mean concave points}).

On the other hand, if we are satisfied we should check whether other
sets of observations agree or conflict with the hypothesis. For instance
we could test our hypothesis on women from different parts of the world.
We could collect data on \textbf{mean area} and
\textbf{mean concave points} that we were not part in
our initial observations.

Sometimes it is not possible to perform another experiment. The next
best thing is to test the hypothesis on the holdout set. The
disadvantage of this is that the data comes from the same place as the
initial training data and so it inherits the same observational bias.

\begin{lstlisting}[numbers=left, language=python, stringstyle=\color{strings}, commentstyle=\color{comments}, morekeywords={import,from,for,in,as,range,if,else,elif}, keywordstyle=\color{keywords}\bfseries, basicstyle=\small]
import sklearn.datasets
import pandas as pd
import feyn
import matplotlib.pyplot as plt

from sklearn.model_selection import train_test_split

breast_cancer = sklearn.datasets.load_breast_cancer()
input_columns = breast_cancer.feature_names

# Load into a pandas dataframe
data = pd.DataFrame(breast_cancer.data, columns=input_columns)
data['target'] = pd.Series(breast_cancer.target)

# Split into train, validation and holdout
train, valid = train_test_split(data, test_size = 0.4, 
                                stratify = data['target'], random_state = 42)
valid, holdo = train_test_split(valid, test_size = 0.5, 
                                stratify = valid['target'], random_state = 42)

# Connecting to QLattice
ql = feyn.QLattice()

# Pose a question to QGraph [*]
qgraph = ql.get_classifier(['mean area', 'mean concave points'], output='target', max_depth=1)

for _ in range(20):
    qgraph.fit(train, threads=4)
    ql.update(qgraph.best())

# Selecting hypothesis [*]
hypo_mean_area_conc_points = qgraph[0]

# Analyse hypothesis [*] --> refer to previous section
\end{lstlisting}

Notice the \textbf{[*]} symbols in the code cell above. They
indicate the points of iteration in the \textit{question-hypothesis}
process. In other words, they represent where we pose questions, extract
hypotheses and analyse them to further refine said questions and get
more robust hypotheses.

Suppose we are satisfied with the hypothesis that the
\textbf{target} variable is a linear function of
\textbf{mean area} and
\textbf{mean concave points}. Let's see how it performs
on the holdout set:

\begin{lstlisting}[numbers=left, language=python, stringstyle=\color{strings}, commentstyle=\color{comments}, morekeywords={import,from,for,in,as,range,if,else,elif}, keywordstyle=\color{keywords}\bfseries, basicstyle=\small]
# Testing hypothesis on unseen data (holdout)
hypo_mean_area_conc_points.plot_roc_curve(holdo)
\end{lstlisting}

\begin{figure}
\centering
\includegraphics{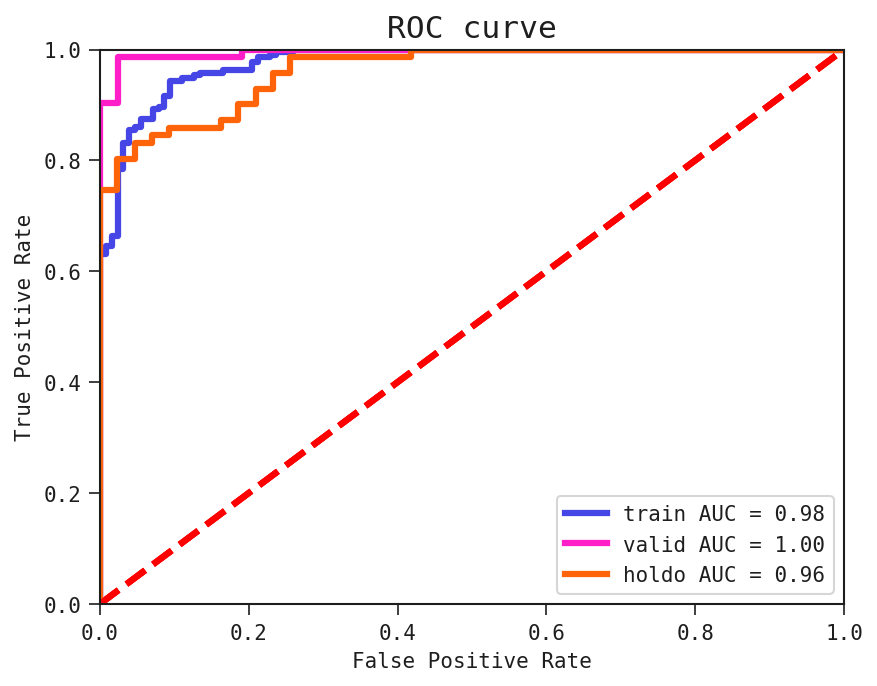}
\caption{ROC curve for train, validation and holdout set}
\label{fig:roc_train_val_holdout}
\end{figure}

From the \textit{AUC score} and the \textit{ROC curve} in Figure \ref{fig:roc_train_val_holdout}, we see
that this simple hypothesis generalises to the holdout set.

If the hypothesis had not generalised to the holdout set, then we are in
trouble. Our holdout set is contaminated, i.e.~the knowledge on this
performance would bias further investigations. Basically the holdout
set became another validation set. Ideally, we should get more unseen
data.

Even though this is a simple data set, hypotheses formed by the \texttt{QLattice} often show this tendency to perform consistently on the holdout set as compared to the train and validation sets. This allows you to have better confidence in the results you get, and know what to expect. Not giving in to excess complexity, having models you can reason about and the mathematical nature of symbolic regression is in part to thank for this.

\section{Conflicts of interest}
All of the authors are employees of either Abzu or Abzu Barcelona.

\newpage
\bibliographystyle{unsrt}
\bibliography{references}

\end{document}